\documentclass[twocolumn]{fairmeta}
\usepackage[most]{tcolorbox}

\usepackage{amsmath}
\usepackage{amssymb}
\usepackage{ulem}

\newcommand{\fref}[1]{Fig.~\ref{#1}}

\newcommand{\tref}[1]{Tab.~\ref{#1}}

\newcommand{\m}[1]{{TF-DP}}

\title{Trace-Focused Diffusion Policy for Multi-Modal Action Disambiguation in Long-Horizon Robotic Manipulation}

\author[1,*]{Yuxuan Hu}
\author[1,*]{Xiangyu Chen}
\author[1]{Chuhao Zhou}
\author[1]{Yuxi Liu}
\author[1]{Gen Li}
\author[1]{Jindou Jia}
\author{Jianfei Yang$^{1,\dagger}$}

\affiliation[1]{MARS Lab, Nanyang Technological University}
\abstract{
Generative model-based policies have shown strong performance in imitation-based robotic manipulation by learning action distributions from demonstrations. 
However, in long-horizon tasks, visually similar observations often recur across execution stages while requiring distinct actions, which leads to ambiguous predictions when policies are conditioned only on instantaneous observations, termed as multi-modal action ambiguity (MA$^2$).
To address this challenge, we propose the Trace-Focused Diffusion Policy (TF-DP), a simple yet effective diffusion-based framework that explicitly conditions action generation on the robot’s execution history. 
TF-DP represents historical motion as an explicit execution trace and projects it into the visual observation space, providing stage-aware context when current observations alone are insufficient. 
In addition, the induced trace-focused field emphasizes task-relevant regions associated with historical motion, improving robustness to background visual disturbances.
We evaluate TF-DP on real-world robotic manipulation tasks exhibiting pronounced multi-modal action ambiguity and visually cluttered conditions. 
Experimental results show that TF-DP improves temporal consistency and robustness, outperforming the vanilla diffusion policy by \textbf{80.56\%} on tasks with multi-modal action ambiguity and by \textbf{86.11\%} under visual disturbances, while maintaining inference efficiency with only a \textbf{6.4\%} runtime increase.
These results demonstrate that execution-trace conditioning offers a scalable and principled approach for robust long-horizon robotic manipulation within a single policy. Project site: \url{https://ntumars.github.io/project/TFDP}
}

\correspondence{Jianfei Yang at \email{jianfei.yang@ntu.edu.sg}}

\begin{document}

\maketitle

\section{Introduction}

Learning from human demonstrations has become a central paradigm for autonomous robotic manipulation, offering an effective way for robots to acquire complex skills.
Through Behavior Cloning (BC)~\cite{bain1995framework, florence2022implicit}, robotic manipulators can imitate expert behaviors without explicit task modeling or reward design.
Recently, generative policies, e.g., diffusion-based and flow-matching-based methods~\cite{chi2025diffusion}, have further advanced this line of work by modeling high-dimensional action distributions through iterative denoising processes. Compared to deterministic policies, generative approaches naturally support multi-modal action prediction, enabling the representation of diverse valid behaviors under the same observation.

\begin{figure*}[tbhp]
    \centering
    \includegraphics[width=\textwidth]
    {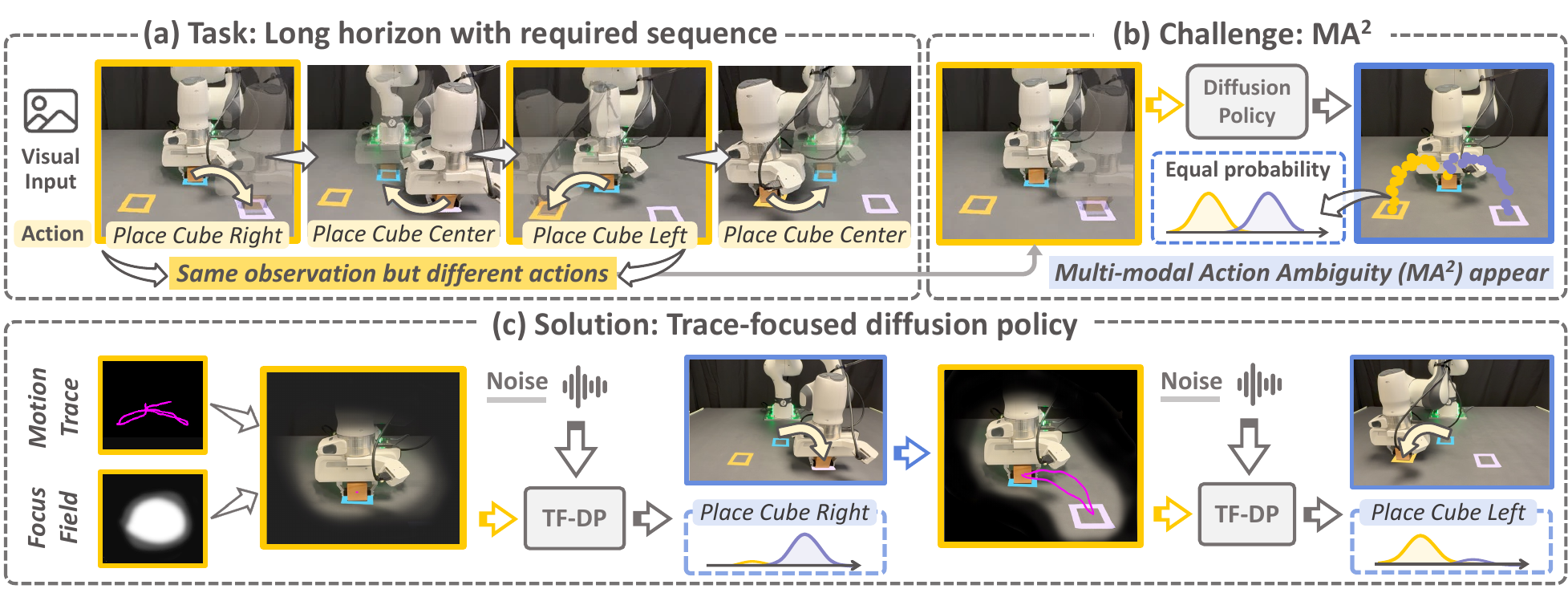}
    \caption{\textbf{Trace-Focused Diffusion Policy for Resolving Multi-modal Action Ambiguity (MA$^2$).} (a) In long-horizon manipulation, visually similar observations map to different actions at different execution stages. (b) This one-to-many mapping causes MA$^2$ for diffusion policies conditioned only on instantaneous observations. (c) TF-DP resolves MA$^2$ by conditioning on explicit motion traces and a trace-focused field, enabling temporally consistent actions. }
    \label{fig:first_page}
    \vspace{-3mm}
\end{figure*}

Despite these advantages, generative policies have an intrinsic issue in maintaining temporal consistency during long-horizon manipulation.
As illustrated in Fig.~\ref{fig:first_page}, action generation in diffusion policies is primarily conditioned on instantaneous observations.
In long-horizon tasks, visually indistinguishable observations often recur at different execution stages (e.g., placing a cube from the center to the right and to the left), inducing a one-to-many observation–action mapping, where the same observation corresponds to multiple valid actions.
Consequently, conditioning only on instantaneous observations can lead to several incompatible actions with comparable likelihoods.
During long-horizon execution, such ambiguity results in unstable action sampling and incorrect temporal ordering, causing the policy to deviate from the intended action sequence. 
This issue is further exacerbated by visual disturbances, which amplify ambiguity in the observation space and lead to compounding errors over time. 
Addressing this limitation is critical for improving the robustness of long-horizon manipulation in real-world deployments.

We refer to this challenge as \textit{multi-modal action ambiguity} (MA$^2$), which fundamentally limits the effectiveness of single-policy diffusion models in long-horizon manipulation when visually similar observations recur across execution stages. 
To avoid MA$^2$, a common workaround is to decompose long-horizon tasks into simpler sequential sub-tasks~\cite{black2025pi, hao2025chd,sun2024hierarchical} using hierarchical architectures or LLM-based planners~\cite{brohan2023can,driess2023palm}, thereby leveraging a series of short-term policies. 
However, such approaches rely on high-level reasoning modules that introduce substantial latency, computational overhead, and reliability issues due to planning errors or hallucinations, making them unsuitable for many real-world settings. 
In practice, a wide range of manipulation scenarios, e.g., fast and precise assembly in manufacturing, still require a single, reactive policy capable of long-horizon execution with low latency and high reliability. 
This highlights the importance of addressing MA$^2$ within a single policy and raises a fundamental question: \textbf{Can MA$^2$ be resolved within a single policy while preserving the efficiency, stability, and reactivity required for real-world long-horizon manipulation?}

Inspired by the \textit{concept of traces} in memory theory~\cite{baddeley1992working} and particle physics~\cite{perkins2000introduction}, where accumulated history shapes future decisions and motions, we develop a simple insight for addressing MA$^2$: incorporating execution history as an explicit trace can provide essential disambiguating context beyond instantaneous observations. 
When visually similar observations recur, differences in the corresponding traces naturally distinguish execution stages, allowing the policy to resolve MA$^2$ and maintain temporal consistency. 
In addition, the trace implicitly emphasizes execution-relevant regions in the observation space, guiding attention~\cite{itti2002model} toward historically salient cues and mitigating the impact of visual disturbances.

% \begin{figure*}[tbhp]
%     \centering
%     \includegraphics[width=\textwidth]{MARS_Lab___Arxiv_Template/figures/Figure_2_arXiv.pdf}
%     \caption{\textbf{The framework of the proposed \m{}}. The historical robot motions are collected to create the motion trace. The proposed Trace-Focused Field is generated from the trace. Then, the trace and focused field are projected to the image space to resolve the MA$^2$ and mitigate the visual disturbance in the background.}
%     \label{fig:fig1}
% \end{figure*}

To this end, we propose the Trace-Focused Diffusion Policy (\m{}), a single-policy framework that explicitly incorporates execution history into diffusion-based action generation. 
\m{} aggregates historical robot motions into a compact execution trace that conditions the policy, enabling effective disambiguation of multi-modal actions when visually similar observations recur. 
As illustrated in~\fref{fig:first_page}, the accumulated trace provides execution-aware context that guides action prediction and promotes temporally consistent behavior over long horizons. 
In addition, the motion trace induces a trace-focused perceptual field that emphasizes task-relevant regions while suppressing irrelevant visual variations, improving robustness under visual disturbances. 
Overall, \m{} demonstrates that long-horizon action ambiguity can be addressed within a single diffusion policy through lightweight history conditioning. 

Our main contributions are summarized as follows:
\begin{itemize}
    \item We identify the MA$^2$ as a fundamental failure mode of diffusion policies in long-horizon robotic manipulation and introduce execution traces as an effective mechanism to resolve it.
    \item We propose the Trace-Focused Diffusion Policy (\m{}), which integrates execution history into diffusion-based action generation to achieve temporally consistent and robust behavior within a single policy.
    \item We validate our approach on challenging real-world manipulation tasks, demonstrating improved action disambiguation and robustness to visual disturbances in long-horizon execution.
\end{itemize}

\section{Related works}
\subsection{Generative Policies for Robot Learning}
Generative model–based policies~\cite{fu2024mobile, chi2025diffusion, zeng2021transporter} have become a widely adopted paradigm for imitation learning~\cite{osa2018algorithmic, hussein2017imitation}.
These methods learn a mapping from visual observations to robot actions by modeling action distributions.
However, policy behavior is heavily dependent on visual inputs.
Even minor changes in scene appearance, background clutter, or lighting conditions can alter the perceived observations, leading to unstable or inconsistent action predictions.
In particular, similar visual observations may admit MA$^2$, making action prediction inherently uncertain under visual-only conditioning.
To address such ambiguity, a growing body of work~\cite{zhang2025language,ke2024diffuseractor} has explored disambiguating action prediction under visually similar observations, though challenges remain in terms of efficiency and robustness for long-horizon tasks.

\subsection{Long-Horizon Manipulation Policies}
In long-horizon robotic manipulation, a common strategy for handling extended temporal dependencies is hierarchical learning, which decomposes complex tasks into ordered sequences of subtasks~\cite{hao2025chd, sun2024hierarchical, ma2024hierarchical, luo2024multistage}.
This formulation effectively tames long-horizon complexity by reducing extended execution into manageable, temporally ordered subtasks.
However, the resulting temporal structure is imposed at the task level through explicit decomposition, rather than emerging from continuous execution.
As a consequence, such approaches are inherently task-specific, relying on carefully designed task decompositions that are costly to construct and difficult to reuse across different manipulation scenarios.

Motivated by the limitations, a substantial body of prior work~\cite{chen2025history, zhou2025mtil, torne2025learning, zheng2024tracevla} has explored incorporating historical information to provide additional temporal context for action prediction in long-horizon tasks.
In practice, temporal context is most commonly incorporated in a minimal form by stacking a fixed number of past observations or actions during policy execution~\cite{chi2025diffusion, zhao2023learning}.
However, temporal stacking is constrained in practice by computational overhead and execution latency, which limit usable history length and weaken the influence of prior actions, resulting in limited effectiveness for resolving action ambiguity.
Another line of research~\cite{guhur2023instruction, black2024pi_0, peng2024learning, torne2025learning} leverages semantic or symbolic context as a form of history, conditioning policies on language instructions, task descriptors, or abstract state representations.
While effective in certain scenarios, these approaches introduce substantially more complex system pipelines, requiring additional models, supervision, or symbolic representations, rather than grounding temporal context in robot dynamics.

\section{Problem definition}
When performing long-horizon tasks, a certain observation can correspond to multiple valid actions. 
As illustrated in Fig.~\ref{fig:first_page}, the actions ``Place Cube Right'' and ``Place Cube Left'' correspond to visually identical observations at different execution stages. Because both actions appear in the demonstration dataset under the same observation, the policies assign comparable likelihoods to them during inference, resulting in ambiguous action predictions.
We define this issue as \emph{multi-modal action ambiguity} (MA$^2$), which should be minimized to enable correct execution order in long-horizon tasks.

Formally, we formulate an imitation-learning dataset as $\mathcal{D} = \{(\mathbf{O}_i, \mathbf{A}_i)\}_{i=1}^{N}$, where $\mathbf{O}_i = \{ o_i^t \}_{t=0}^{T}$ denotes a temporal sequence of observations and $\mathbf{A}_i = \{ a_i^t \}_{t=0}^{T}\in \mathbb{R}^{T \times 3}$ represents the corresponding expert actions. For clarity, we omit the data sample index $i$ in the following paper. At each time step $t$, the observation $o_t = \left\{ I_t^{\mathrm{global}},\, I_t^{\mathrm{side}},\, I_t^{\mathrm{wrist}},\, p_t^{\mathrm{ee}} \right\}$ combines global-, side-, wrist-view camera images $I_t^{\mathrm{global}}, I_t^{\mathrm{side}}, I_t^{\mathrm{wrist}} \in \mathbb{R}^{128 \times 128}$ and end-effector position $p_t^{\mathrm{ee}}= (x_t, y_t, z_t) \in \mathbb{R}^3$.

\begin{figure*}[tbhp]
    \centering
    \includegraphics[width=\textwidth]{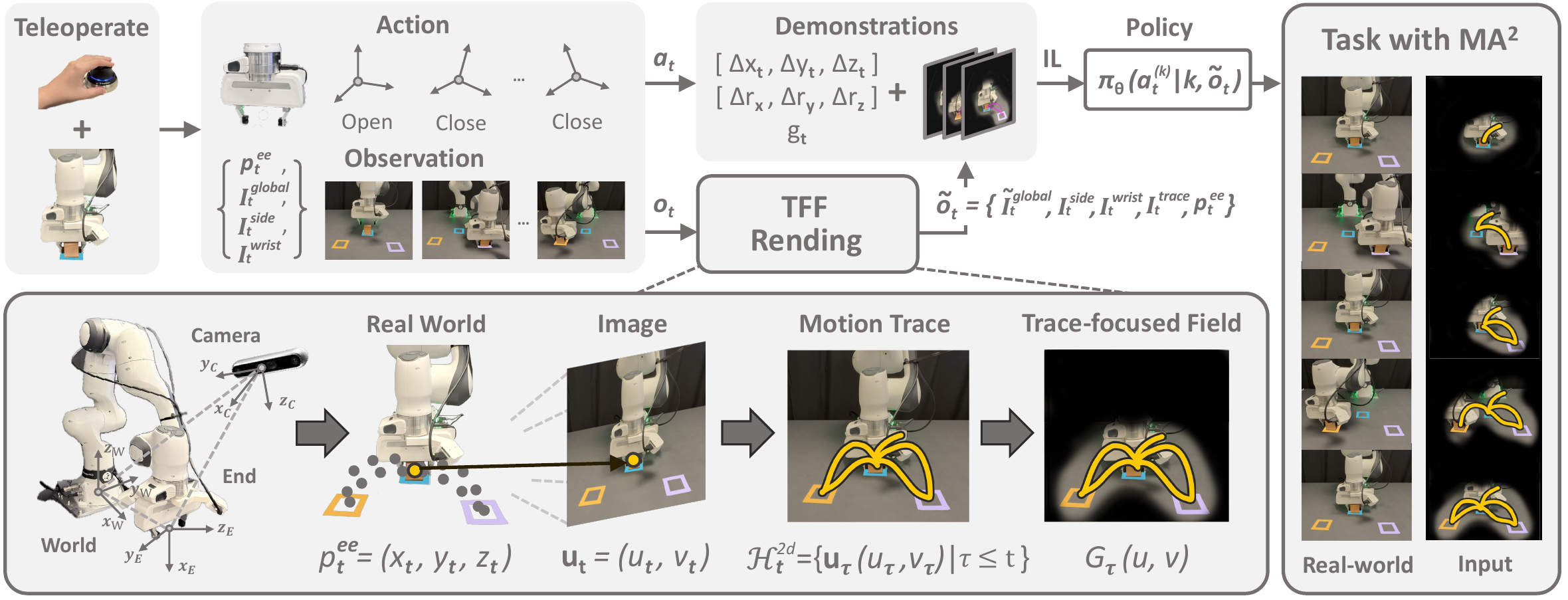}
    \caption{\textbf{The framework of the proposed \m{}}. The historical robot motions are collected to create the motion trace. The proposed Trace-Focused Field is generated from the trace. Then, the trace and focused field are projected to the image space to resolve the MA$^2$ and mitigate the visual disturbance in the background.}
    \label{fig:framework}
     % \vspace{-4mm}
\end{figure*}

The goal is to learn a robotic policy $\pi_\theta$ to reproduce expert behaviors. 
In standard behavior cloning, the policy models the conditional distribution $\pi_\theta(a_t|o_t)$ based on current observations. However, when observations at different timesteps are highly similar, the MA$^2$ problem emerges:
% \begin{equation}
% \pi_\theta(\cdot \mid o_m) \approx \pi_\theta(\cdot \mid o_n),
%  \text{if } o_m \approx o_n,
%  \text{while } a_m^\ast \neq a_n^\ast .
% \label{eq:multi_modal_action_ambiguity}
% \end{equation}
% \begin{equation}
% \begin{aligned}
% \pi_\theta(\cdot \mid o_m) \approx \pi_\theta(\cdot \mid o_n), \\
% \text{if } o_m \approx o_n, \text{ while } a_m^* \neq a_n^* .
% \end{aligned}
% \end{equation}
\begin{equation}
\begin{gathered}
\pi_\theta(\cdot \mid o_m) \approx \pi_\theta(\cdot \mid o_n), \\
\text{if } o_m \approx o_n, \text{ while } a_m^* \neq a_n^* .
\end{gathered}
\end{equation}
This problem arises from the training process, where the observation–action relationship is inherently a one-to-many mapping: the visually similar observations are associated with multiple valid actions across different execution stages.
As a result, the policy learns to assign high probability to several action modes conditioned on the similar observations. 
During inference, this leads to comparable likelihoods over several actions, giving rise to MA$^2$.
This ambiguity hinders the policy from following the correct action sequence, thereby degrading performance in long-horizon tasks.
In this work, we address this issue by leveraging the historical traces $\mathcal{H}_t$ at step $t$ to steer the policy in inferring the current task execution stage and dynamically adjusting its predicted action distribution. The policy can then sample stage-aware actions, thereby maintaining the correct action order in long-horizon tasks.
% \vspace{-0.73cm}

\section{Methodology}
As shown in~\fref{fig:framework}, we propose the Trace-Focused Diffusion Policy (\m{}), which integrates historical traces via Trace-Focused Field (TFF) Rendering to reduce MA$^2$. 
Given the observation $o_t = \{ I_t^\mathrm{global}, I_t^\mathrm{side}, I_t^\mathrm{wrist}, p_t^\mathrm{ee}\}$, historical end-effector trace $\mathcal{H}_t = \left\{ p_\tau^{\mathrm{ee}} = (x_\tau, y_\tau, z_\tau) \mid \tau \le t \right\}$ and action $a_t$ at timestep $t$, the TFF module first projects $\mathcal{H}_t$ from 3D robot space into the 2D global camera space, getting the trace image $I^{trace}_t$. 
A trace-focused field is then rendered over $I_t^{\mathrm{global}}$ based on the projected 2D trace, producing an enhanced global view $\tilde{I}_t^{\mathrm{global}}$ that explicitly encodes historical execution information.
By conditioning the denoising process on execution-aware observations $\tilde{o}_t = \{\tilde{I}_t^{\mathrm{global}}, I_t^{\mathrm{side}}, I_t^{\mathrm{wrist}}, I^{trace}_t, p_t^{\mathrm{ee}}\}$, our \m{} learns to associate visually similar observations with distinct actions corresponding to different execution stages in long-horizon tasks. Consequently, manipulation tasks that require preserving temporal order, such as sequentially striking keyboard keys, can be essentially addressed by our \m{}.

\subsection{Trace-Focus Diffusion Policy}
\label{subsec:TF-DP}
To model historical traces, existing methods typically incorporate previous action tokens into the input of the diffusion policy~\cite{chi2025diffusion}. 
However, as task execution proceeds, the number of historical tokens in the input grows as well, substantially increasing the computational cost of the denoising process. 
Furthermore, we reveal in~\tref{tab:across_task} and~\fref{fig:eff_analysis} that adding historical actions from several prior steps (DP-HistAct) can perform even worse than the original diffusion policy (DP) with a significant increase in usage of GPU memory. 
When observing a moving object, human vision primarily attends to its trajectory and the immediate surroundings~\cite{johansson1973visual}. 
Motivated by this fact, we directly integrate historical motion information into the global visual observation $I_t^{\mathrm{global}}$ using Trace-Focused Field (TFF) rendering. The enhanced $\tilde{I}_t^{\mathrm{global}}$ is both efficient and effective: it leverages one image to incrementally record the entire motion throughout task execution while simultaneously emphasizing motion traces and suppressing task-irrelevant visual clutter. The acquisition and rendering of historical motion traces are detailed in the following subsections.

\subsubsection{Motion Trace Acquisition}
\label{subsubsec: Motion Trace}
To integrate the history motion information for the action prediction without increasing the number of neural parameters within the action prediction model, we choose to project the trace onto the global observation as one guidance.
We consider three coordinate frames: the world frame $\{W\}$, the end-effector frame $\{E\}$, and the global camera frame $\{C\}$. For simplicity, we treat the world frame as coincident with the robot base frame.
At time step $t$, the historical motion trace $\mathcal{H}^W_t = \left\{ p_\tau^{\mathrm{ee}} \mid \tau \le t \right\}$ represents the end-effector position in the$\{W\}$ frame.
The $\mathcal{H}_t$ is first transformed into the $\{C\}$ frame to get $\mathcal{H}^C_t=T^C_W\mathcal{H}_t$, according to the transformation matrix $T^C_W$, obtained from the camera calibration~\cite{zhang2002flexible}.
We project $\mathcal{H}^C_t$ onto the global image $I_t^{\mathrm{global}}$ instead of encoding $\mathcal{H}^C_t$ as an additional network input. 
For each 3D end-effector position $p^{ee}_t = (x_t, y_t, z_t)^\top\in\mathcal{H}^C_t$, the corresponding 2D image-space coordinate $\mathbf{u}_t= (u_t, v_t) \in \mathbb{R}^2$ is obtained via standard perspective projection:
\begin{equation}
\begin{aligned}
\begin{bmatrix}
\tilde{u}_t \\ \tilde{v}_t \\ \tilde{z}_t
\end{bmatrix}
&=
\begin{bmatrix}
f_x & 0 & c_x \\
0 & f_y & c_y \\
0 & 0 & 1
\end{bmatrix}
\begin{bmatrix}
x_t \\ y_t \\ z_t
\end{bmatrix}, \\
(u_t, v_t)
&=
\left(
\frac{\tilde{u}_t}{\tilde{z}_t},
\frac{\tilde{v}_t}{\tilde{z}_t}
\right).
\end{aligned}
\end{equation}
Applying this projection over the execution horizon yields a set of image-space end-effector trace points in 2D space,
\begin{equation}
\mathcal{H}^{2d}_t =
\{ \mathbf{u}_\tau = (u_\tau, v_\tau) \in \mathbb{R}^2 \mid \tau \leq t \},
\end{equation}
which can be used to paint the trace on the global observation. 
Thus, the trace can be projected to the image to get the $I_t^{trace}$. 
This design choice avoids significantly increasing the input dimensionality and preserves architectural simplicity, while exploiting the strong representational capacity of modern image encoders to jointly model visual content and execution history from the augmented image.

\subsubsection{Trace-Focused Field Rendering}
\label{subsubsec:Trace-Focused Field} 
The motion trace in the global camera view $\mathcal{H}^{2d}_t$ consists of discrete trajectory points associated with the task. 
As discussed in~\ref{subsec:TF-DP}, humans attend to both the trajectory positions and their surrounding regions when observing a moving object. To this end, we propose Trace-Focused Field (TFF) rendering to extend the discrete motion trace $\mathcal{H}^{2d}_t$ into a continuous energy field $\mathcal{F}^{2d}_t$ over the global observation space. 
Specifically, a dense Gaussian response map is calculated for each discrete point $\mathbf{u}_{\tau} \in \mathcal{H}^C_t$: 
\begin{equation}
G_{\tau}(u,v)
=
\exp\!\left(
- \frac{(u - u_{\tau})^2 + (v - v_{\tau})^2}{2\sigma^2}
\right) \in \mathbb{R}^{H \times W},
\end{equation}
where $(u,v)$ denotes a pixel index on the image plane with
$u \in {1,\dots,W}$ and $v \in {1,\dots,H}$. We then render the energy field $\mathcal{F}^C_t$ by summation of all dense Gaussian maps:
\begin{equation}
\mathcal{F}_t^{2d} = \sum_{\tau=1}^{t} G_\tau(u,v) \in \mathbb{R}^{H \times W}, 
\end{equation}
Consequently, the execution-aware global observation $\tilde{I}_t^\mathrm{global}$ can be achieved via:
\begin{equation}
\tilde{I}_t^\mathrm{global} = I_t^\mathrm{global} \odot \mathcal{F}^{2d}_t, 
\end{equation}
In this way, $\tilde{I}_t^\mathrm{global}$ can encode historical motion traces and highlight task-relevant areas at different execution stages. 
\begin{figure*}[t]
\centering
    \includegraphics[width=\linewidth]{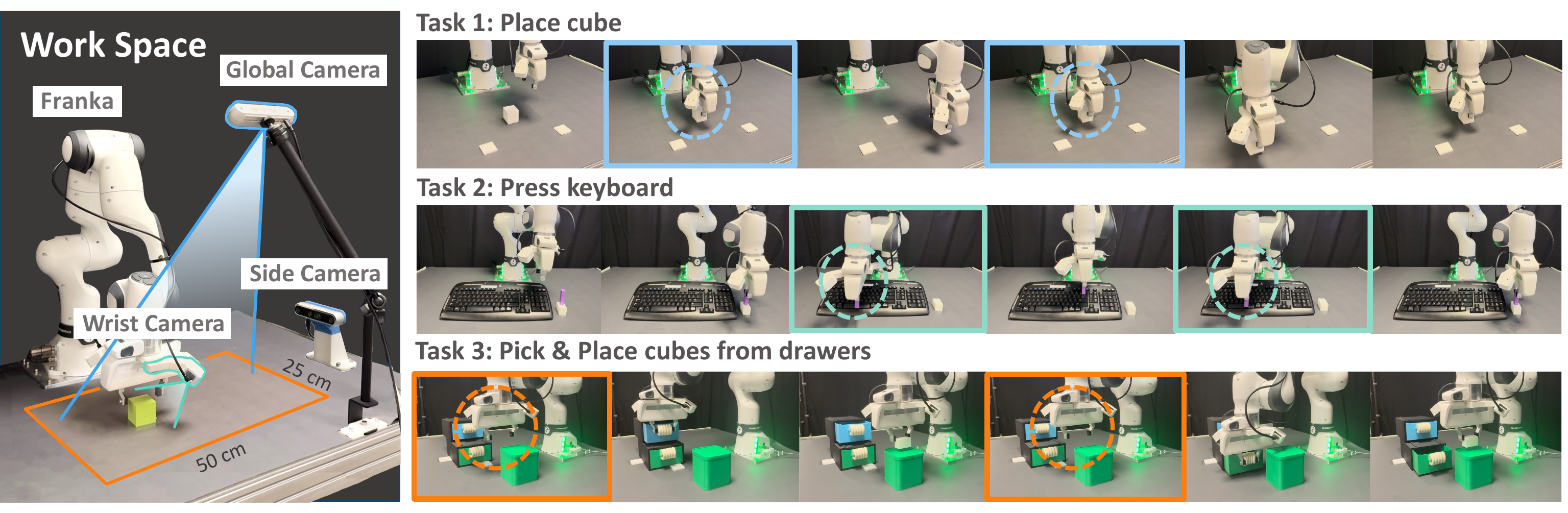}
    \caption{\textbf{The experimental setup and the evaluation tasks. }In the workspace, Franka Research 3 is used as the robotic manipulation platform with one wrist camera, one side camera, and one top camera. Three evaluation tasks, including Place cube, Press keyboard, and Pick \& Place cubes from drawers, are selected since they all have MA$^2$. The multicolored dashed circle represents the scene when the robot has the same observation but requires different action choices.}
    \label{fig:exp:setup}
    \vspace{-4mm}
\end{figure*}

\subsection{\m{} Training \& Inference}
\label{subsec: TF-DP Training}
Once the $\tilde{I}_t^\mathrm{global}$ is achieved, the objective of \m{} training is to learn a mapping from current observations augmented with accumulated execution traces to the next action, ensuring that action decisions remain in correct temporal order under recurring or visually similar observations.
To this end, we adopt a diffusion-based imitation learning framework and execute the denoising process conditioned on execution-aware observation $\tilde{o}_t = \{\tilde{I}_t^{\mathrm{global}}, I_t^{\mathrm{side}}, I_t^{\mathrm{wrist}}, I^{trace}_t, p_t^{\mathrm{ee}}\}$.
Particularly, we adopt the Resnet18~\cite{he2016deep} and an MLP layer to encode vision-based observations and the end-effector position, respectively. For the $t$-th expert action $a_t$, Gaussian noise is then injected according to the forward diffusion process:
\begin{equation}
    a_t^{(k)} = \sqrt{\alpha_k}\, a_t + \sqrt{1-\alpha_k}\, \epsilon,
\quad \epsilon \sim \mathcal{N}(0, I),
\end{equation}
where $k$ denotes the diffusion timestep and $\{\alpha_k\}$ is a predefined schedule controlling the action-to-noise ratio at step $k$.
The $\pi_\theta$ is trained to predict the injected noise conditioned on both the diffusion timestep and the execution-aware observation by minimizing the following objective:
\begin{equation}
\mathcal{L}_{DP} =
\left\| \epsilon - \pi_\theta\!\left(a_t^{(k)} \mid k, \, \tilde{o}_t\right) \right\|^2.
\end{equation}
This encourages the policy to recover the action distribution by referring to the historical execution trace.
As a result, \m{} learns to generate temporally correct actions as the task proceeds, without incurring extra computational overhead.

During training, the Trace-Focused Field is generated in an open-loop manner using the complete demonstration trajectory, which allows the field at each timestep to be rendered directly.
In contrast, TF-DP operates in a closed-loop manner during inference.
Therefore, the Trace-Focused Field is rendered incrementally at each timestep based on the accumulated execution trace observed online: $\mathcal{H}^C_{t+1} = \mathcal{H}^C_{t} \cup p_{t+1}^{ee}$, where $p_{t+1}^{ee}$ denotes the end-effector position at timestep $t+1$.

\section{Experiments}
The experimental evaluation focuses on long-horizon robotic manipulation tasks with MA$^2$ issues. 
We compare our approach against baselines to systematically evaluate the effectiveness of~\m{}.
The experimental design aims to answer the following research questions:
\begin{itemize}
    \item \textit{Q1:} Can the proposed \m{} effectively resolve the MA$^2$ problem in lone-horizon tasks?
    \item \textit{Q2:} Does the proposed trace-focused mechanism improve robustness to background visual disturbances by emphasizing task-relevant regions?
    \item \textit{Q3:} Does the proposed TF-DP address the MA$^2$ issue and overcome the visual disturbances in a computationally efficient manner?
    \item \textit{Q4:} Is the proposed mechanism compatible with and effective across different generative policy formulations?
\end{itemize}

\subsection{Experimental Setup \& Evaluation Tasks}
We construct a real-world robotic manipulation platform to achieve three challenging tasks that exhibit obvious MA$^2$ problems. 
These tasks require the robot to generate distinct action sequences under similar visual observations, posing significant challenges for imitation-based policies. 
The overall workspace configuration and representative task snapshots are illustrated in~\fref{fig:exp:setup}.

\subsubsection{Workspace}
All experiments are conducted using a Franka Emika Research 3 (FR3) robotic arm within a structured tabletop workspace.
The workspace spans approximately 50 cm × 25 cm, within which all task-relevant objects are placed and manipulated.
The scene is observed by three RGB-D Intel RealSense D455 cameras: a fixed overhead camera providing a global view of the workspace, a side-view camera offering an assistant perspective to capture lateral interactions and object states, and a wrist-mounted camera attached to the end-effector that supplies an egocentric view during manipulation.
Policy training and inference are performed on a workstation equipped with an NVIDIA RTX 5080 GPU (16 GB) and 128 GB of system memory.
All human demonstrations are collected using a 3Dconnexion SpaceMouse via the HIL-SERL teleoperation framework~\cite{luo2025precise}.

\begin{figure*}[t]
\centering
    \includegraphics[width=\linewidth]{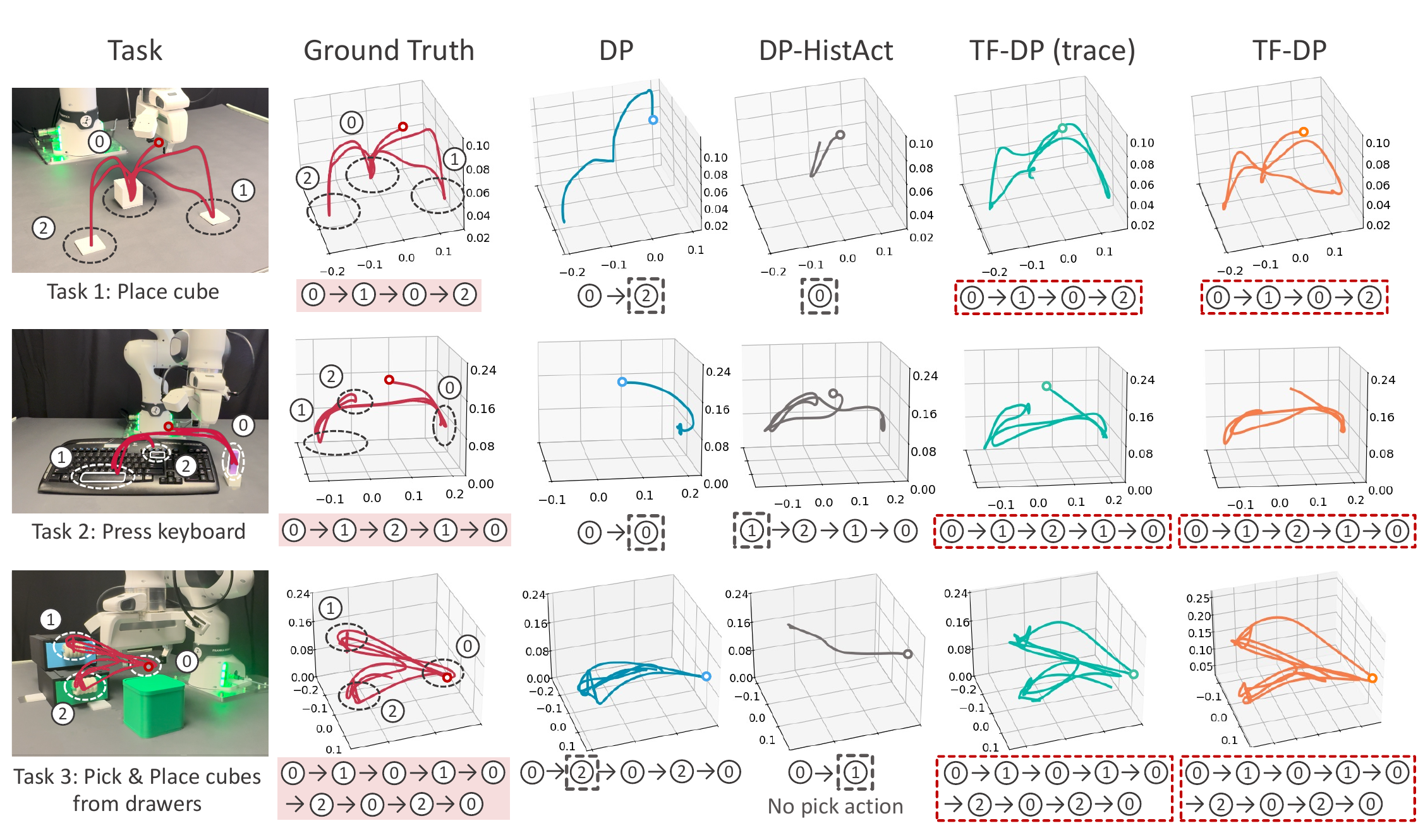}
    \caption{\textbf{Execution trajectory comparisons under tasks with the MA$^2$ problem.} We visualize the representative trajectories for the selected tasks (Place cube, Press keyboard, and Pick \& Place cubes from drawers).
    From left to right: task setup with sequence information, ground-truth trajectory from human demonstrations, diffusion policy (DP), DP conditioned on past actions (DP-HistAct), \m{} with only trace, and \m{}. DP and DP-HistAct fail to follow the correct execution order, while \m{} variants produce temporally consistent trajectories that match the demonstrated action sequences.}
    \label{fig:exp:traj}
\end{figure*}

\subsubsection{Evaluation Tasks}
Since there is no established simulation benchmark that focuses on the MA$^2$ problem in long-horizon manipulation, and constructing such tasks in simulation would require substantial additional engineering, we conduct our evaluation directly in a real-world robotic environment.
We evaluate \m{} on three real-world manipulation tasks that exhibit increasing levels of complexity and horizon length, all characterized by inherent action ambiguity. 
\fref{fig:exp:traj} shows representative execution trajectories for 
Task~1 (Place cube), Task~2 (Press keyboard), and Task~3 (Pick \& Place cubes from drawers). 
The colored dashed circles indicate visually similar observations that occur at different execution stages but require different subsequent actions, resulting in MA$^2$. For Task 1 and Task 2, we collect 30 demonstrations for training policies, while for Task 3, which has a longer execution horizon, we collect 50 demonstrations.
\begin{itemize}
    \item \textit{Task 1 (Basic):} 
    The robot alternates cube placements by moving the cube from the center to the right location, returning to the center, and then placing it from the center to the left.
    At the visually similar center state, the correct next action is bi-modal, moving either left or right, depending on the execution history.

    \item \textit{Task 2 (Complex):} 
    The robot performs a realistic keyboard interaction sequence by pressing keys in order, moving from the space key to the delete key, and then returning from the space key to the home position.
    At the visually similar space-key state, the next action is bi-modal, either pressing the delete key or moving back to the home position, depending on the execution history.

    \item \textit{Task 3 (Long-horizon):} 
    The robot sequentially opens two drawers, retrieves an object from each drawer, discards it into a bin, closes the drawer, and returns to the home position.
    At the visually similar home state after completing the first drawer, the next action becomes ambiguous, either repeating the first sequence or initiating the second, making this a long-horizon task with complex action dependencies and strong stability requirements.
\end{itemize}

In the following experiments, we evaluate each task over 12 independent trials under identical conditions and use the task success rate as the performance metric for all methods.

\begin{figure}[t]
    \centering
    \begin{subfigure}[t]{0.45\linewidth}
        \centering
        \includegraphics[width=\linewidth]{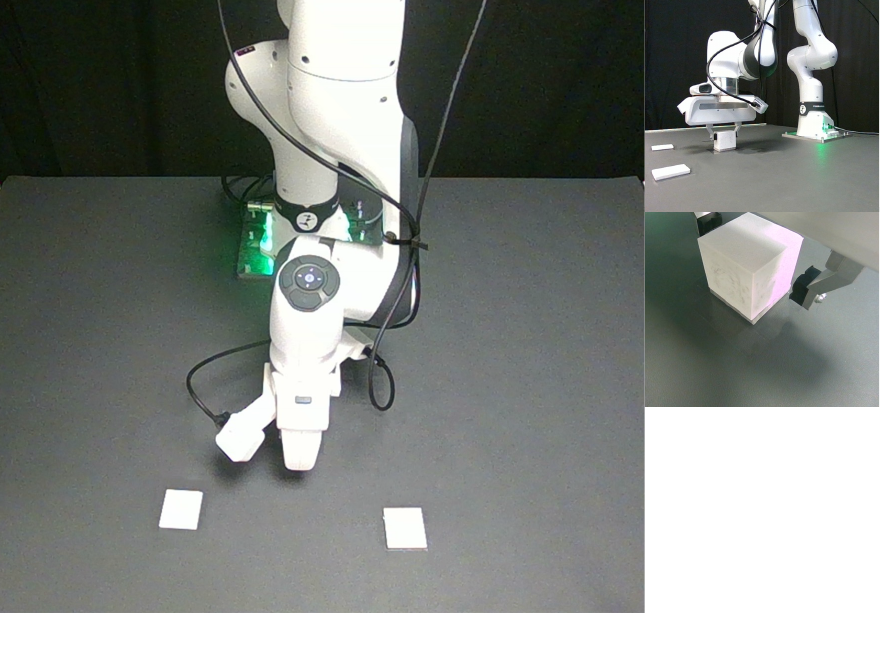}
        \vspace{-1.5em}
        \caption{DP}
    \end{subfigure}
    % \hfill
    % \hspace{0.04\linewidth}
    \begin{subfigure}[t]{0.43\linewidth}
        \centering
        \includegraphics[width=\linewidth]{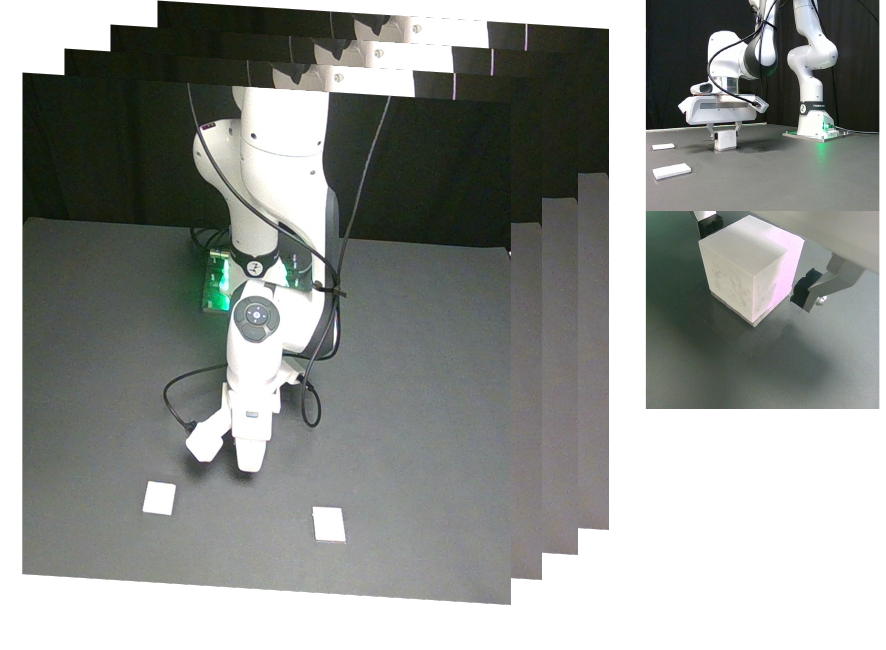}
        \vspace{-1.5em}
        \caption{DP-HistAct}
    \end{subfigure}
    \vspace{0.9em}
    % \hfill 
    \\
    % \hfill
    \begin{subfigure}[t]{0.45\linewidth}
        \centering
        \includegraphics[width=\linewidth]{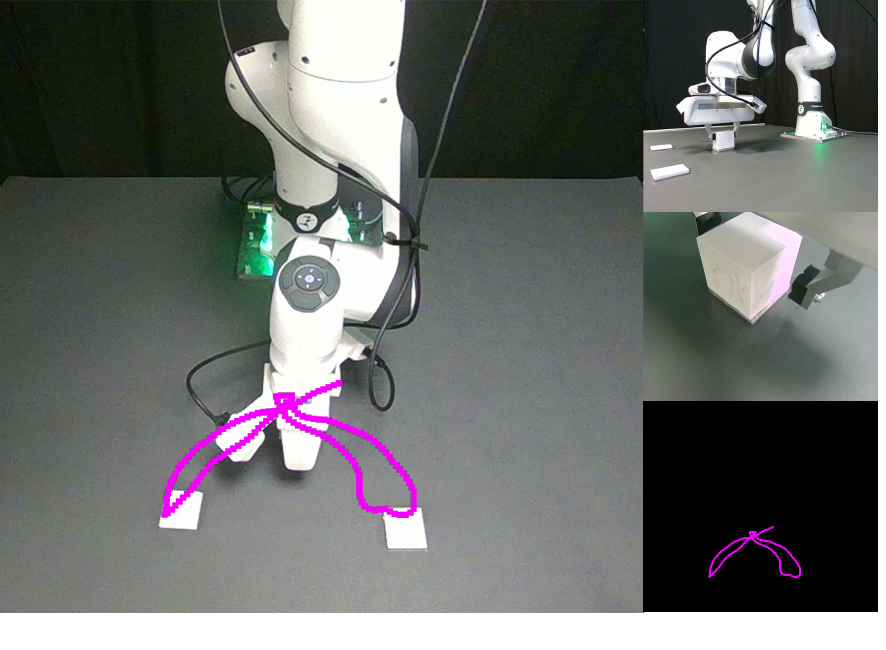}
        \vspace{-1.5em}
        \caption{TF-DP (trace)}
    \end{subfigure}
    % \hspace{0.04\linewidth}
    \begin{subfigure}[t]{0.45\linewidth}
        \centering
        \includegraphics[width=\linewidth]{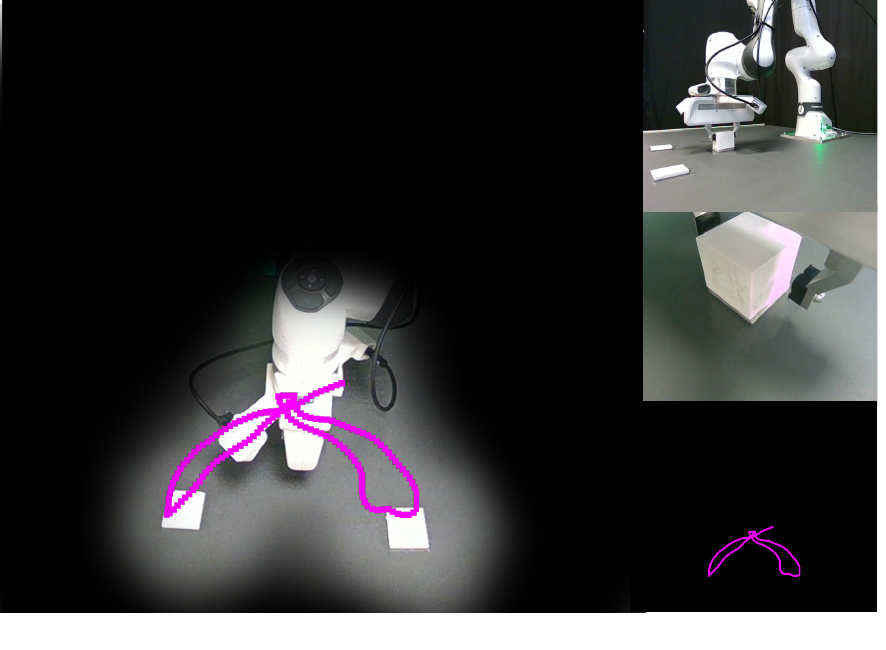}
        \vspace{-1.5em}
        \caption{TF-DP}
    \end{subfigure}
    \caption{\textbf{Visualization of input representations used by different methods.}
(a) DP, Diffusion Policy; (b) DP-HistAct, DP with action history conditioning; (c) TF-DP (trace), TF-DP with execution trace only; and (d) TF-DP, full TF-DP incorporating both execution traces and the trace-focused field.}
    \label{fig:exp:baselines}
    \vspace{-4mm}
\end{figure}

\subsection{Methods for Comparison}
The selected methods differ primarily in their input observation representations, as visualized in~\fref{fig:exp:baselines}.
\subsubsection{Baseline Methods}
Two baseline policies, original DP and DP with additional history observations (DP-HistAct), are selected for experimental comparisons. 
As illustrated in~\fref{fig:exp:baselines}~(a, b), DP predicts actions solely based on the current observations, and DP-HistAct conditions the diffusion process on a fixed window of the past 8 observations, providing temporal historical context.

\subsubsection{Proposed Methods}
Two variants of the proposed \m{}, \m{} (trace) and full~\m{} are selected for evaluations.
As shown in~\fref{fig:exp:baselines}~(c, d), \m{} (trace) projects the executed motion history as an explicit motion trace on the image space, and the full \m{}, further incorporates a trace-focused field that emphasizes the motion-relevant areas. The comparison between \m{} (trace) and the full \m{} constitutes an ablation study that isolates the impact of the explicit motion trace and the trace-focused field on policy performance.

\subsection{Experimental Results}
We conduct three sets of experiments to systematically answer the research questions outlined above.

\noindent\textbf{\textit{(1) Can the proposed \m{} effectively resolve the MA$^2$ problem in lone-horizon manipulation tasks?}}\par
To answer this question, we evaluate all selected methods on three manipulation tasks with MA$^2$ problems. 
As shown in ~\fref{fig:exp:traj}, the original Diffusion Policy (DP) suffers from severe MA$^2$, resulting in trajectories that violate the demonstrated execution order and fail to complete the task. 
Notably, conditioning DP on history actions (DP-HistAct) directly fails the task, producing inconsistent trajectories that diverge from the intended targets. 
In contrast, both \m{} (trace) and the full \m{} faithfully imitate the demonstrated trajectory structure and execute the task in the correct temporal order. 
These results demonstrate that the proposed trace-related information effectively resolves the MA$^2$ problem when facing visually similar observations, enabling reliable long-horizon execution.

\begin{table}[t]
    \centering
    \caption{Task success rates (\%) across the three real-world manipulation tasks with the MA$^2$ problem.}
    \label{tab:across_task}
    \setlength{\tabcolsep}{2pt}
    \renewcommand{\arraystretch}{1.25}
    \begin{tabular}{lcccc}
        \toprule
        \textbf{Method} 
        & \textbf{Task 1} 
        & \textbf{Task 2} 
        & \textbf{Task 3} 
        & \textbf{Average} \\
        \midrule
        DP~           & 16.67\% & 8.33\%  & 8.33\%  & 11.11\% \\
        DP-HistAct~    & 8.33\%  & 16.67\%  & 0.00\%  & 8.33\%  \\
        \m{} (trace)    & \uline{75.00\%} & \uline{75.00\%} & \uline{66.67\%} & \uline{72.22\%} \\
        \m{} & \textbf{91.67\%} & \textbf{100.00\%} & \textbf{83.33\%} & \textbf{91.67\%} \\
        \bottomrule
    \end{tabular}
    \label{tab:exp:real_world}
    \vspace{1mm}
\end{table}

Furthermore, \tref{tab:exp:real_world} reports the task success rates across three real-world manipulation tasks characterized by the MA$^2$ problem.
The standard Diffusion Policy (DP) exhibits consistently low performance across all tasks (11.11\% on average), particularly in scenarios requiring correct action sequencing under visually similar observations, highlighting its inability to resolve the MA$^2$ problem from instantaneous inputs alone.
DP-HistAct presents no significant improvement (8.33\% average success) compared with DP, indicating that naively conditioning on the past several actions is insufficient to overcome the MA$^2$ problem for stable long-horizon manipulation.
In contrast, \m{} (trace) shows an average success rate of 72.22\% on Tasks~1-3, demonstrating the effectiveness of explicitly incorporating execution history as trace.
These gains are consistently observed across basic, complex, and long-horizon tasks, underscoring the general applicability of the proposed approach.
The full \m{} further improves performance by introducing the focus-trace field, achieving the best overall results with an average success rate of 91.67\%, which confirms the benefit of leveraging global execution traces and spatially focused guidance for stable action generation.

\begin{figure}[t]
    \centering
    \begin{subfigure}[t]{0.31\linewidth}
        \centering
        \includegraphics[width=\linewidth]{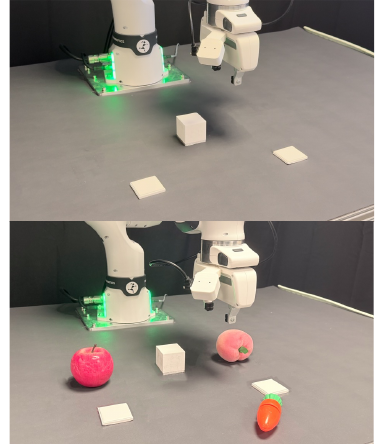}
        \caption{Task 1}
    \end{subfigure}
    % \hfill
    \begin{subfigure}[t]{0.31\linewidth}
        \centering
        \includegraphics[width=\linewidth]{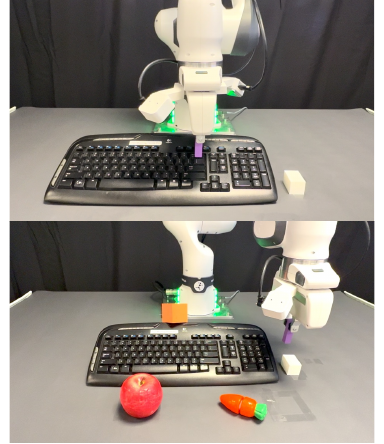}
        \caption{Task 2}
    \end{subfigure}
    % \hfill
    \begin{subfigure}[t]{0.31\linewidth}
        \centering
        \includegraphics[width=\linewidth]{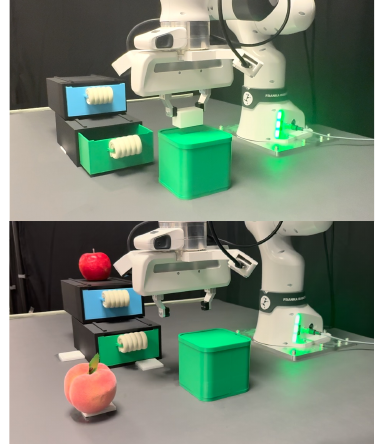}
        \caption{Task 3}
    \end{subfigure}
    \caption{\textbf{Evaluation under visual disturbances.} We introduce additional distractors and background clutter during execution for three tasks: (a) Place cube, (b) Press keyboard, and (c) Pick \& Place cubes from drawers, to assess policy robustness under visually perturbed environments. }
    \label{fig:exp:visual_dis}
\end{figure}

\begin{table}[t]
    \centering
    \caption{Task success rates (\%) across the three real-world tasks under the visual disturbance in the background.}
    \label{tab:exp:visual_disturbance}
    \setlength{\tabcolsep}{3pt}
    \renewcommand{\arraystretch}{1.25}
    \begin{tabular}{lcccc}
        \toprule
        \textbf{Method} 
        & \textbf{Task 1} 
        & \textbf{Task 2} 
        & \textbf{Task 3} 
        & \textbf{Average} \\
        \midrule
        DP    & 0.00\%     & 0.00\%     & 8.33\%    & 2.78\%      \\
        \m{} (trace) & \uline{50.00\%} & \uline{66.67\%} & \uline{58.33\%} & \uline{58.33\%} \\
        \m{} & \textbf{91.67\%} & \textbf{91.67\%} & \textbf{83.33\%} & \textbf{88.89\%} \\
        \bottomrule
    \end{tabular}
    \vspace{-4mm}
\end{table}

\noindent\textbf{\textit{(2) Does the proposed trace-focused mechanism improve robustness to background visual disturbances by emphasizing task-relevant regions?}}\par

To answer this question, we add several distractors in the original task environments, as shown in~\fref{fig:exp:visual_dis}. 
DP, \m{} (trace), and the full \m{} are selected for comparison to isolate the contribution of each module in mitigating visual disturbances.
Each task is evaluated over 12 independent trials, and the success rate is used as the evaluation metric.
The quantitative results are reported in~\tref{tab:exp:visual_disturbance}.

As shown in the table, the original DP fails almost entirely under background disturbances, achieving success rates of 0.00\%, 0.00\%, and 8.33\% on the three tasks, respectively.
In contrast, incorporating the motion trace leads to substantial improvements across all tasks, with success rates increasing to 50.00\%, 66.67\%, and 58.33\%.
On average, this corresponds to an absolute improvement of \textbf{55.55\%} over the original DP, demonstrating that history trace significantly enhances robustness under visually ambiguous and disturbed observations.

Introducing the trace-focused field further improves performance consistently across tasks.
Compared to \m{} (trace), the full \m{} achieves additional gains of 41.67\%, 25.00\%, and 25.00\% on Tasks 1-3, respectively, resulting in a higher and more uniform success rate of 88.89\% on average.
These consistent improvements indicate that the trace-focused field effectively suppresses background-induced noise and guides the policy toward task-relevant regions, yielding more stable and reliable execution in cluttered visual environments.

\noindent\textbf{\textit{(3) Does the proposed TF-DP address the MA$^2$ issue and overcome the visual disturbances in a computationally efficient manner?}}\par
\begin{figure}
    \centering
    \includegraphics[width=\linewidth]{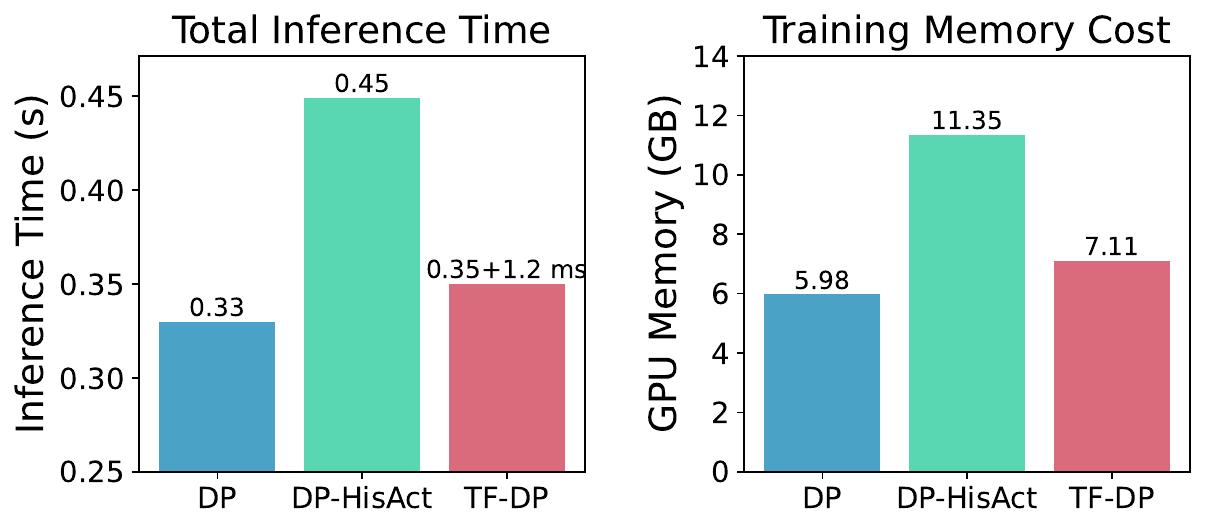}
    \caption{\textbf{Efficiency comparison of the policies.} Left: Inference time, with negligible extra network prediction time (0.02\,s) and the trace-focused field rendering time ($\approx$1.2\,ms). Right: Training memory cost of the compared policies.}
    \label{fig:eff_analysis}
    \vspace{-4mm}
\end{figure}

To answer this question, we measure the per-step inference time and training-time GPU memory consumption of three methods: the original diffusion policy (DP), DP with history actions as additional inputs (DP-HistAct), and the proposed TF-DP.
As shown in~\fref{fig:eff_analysis}, TF-DP incurs only a negligible inference overhead compared to DP, introducing an additional 0.02\,s action chunk (+6.4\%) prediction time with 1.2\,ms trace-focused field rendering time.
Because TF-DP augments all the history motion traces in image space instead of concatenating the history observation and action, its GPU memory consumption increases by about 18\% than that of DP.
In contrast, DP-HisAct substantially increases computational cost, leading to a 36\% increase in inference time and a 107\% increase in training memory usage.
These results indicate that projecting execution traces onto the global observation provides a computational efficient mean of incorporating execution history to overcome the MA$^2$ issue in long-horizon manipulation tasks.

\noindent\textbf{\textit{(4) Is the proposed mechanism compatible with and effective across different generative policy formulations?}}\par
To verify that the proposed Trace-Focused module is not specific to a particular generative formulation, we evaluate its effectiveness when integrated with other representative denoising mechanisms: DDIM~\cite{song2020denoising}, and Flow Matching (FM)~\cite{lipman2022flow}.
In DDIM, the forward diffusion is discretized into 100 noise-adding steps, while sampling is performed with 10 denoising steps. For flow matching, we employ 10 steps for both forward and backward integration.
For each mechanism, we compare the corresponding baseline policy with and without the TF module on Task 3, while keeping the network architecture, conditioning inputs, training data, and optimization settings identical. 
The quantitative results are summarized in~\tref{tab:exp:q4}. As we can see from the table, FM and DDIM completely fail under the evaluated setting, achieving 0.00\% success rate, indicating that policies with these denoising mechanisms also struggle to resolve the MA$^2$ problem without the trace-focused module.
In contrast, incorporating the TF module leads to clear performance improvements for both mechanisms, raising the success rate to 58.33\% for FM and 66.67\% for DDIM.
These results suggest that the proposed TF module is not specific to a particular denoising formulation, but can be effectively combined with different denoising mechanisms to alleviate the MA$^2$ issue in long-horizon manipulation.

In summary, the proposed Trace-Focused DP can effectively overcome the MA$^2$ problem by explicitly integrating the motion trace. 
Besides, the proposed trace-focused field can greatly alleviate the influence of visual disturbance, keeping robustness to action in the environment with multi-distractors.
Experimental results further demonstrate that the TF-DP use an efficient way to solve the MA$^2$ problem and mitigate the influence of the visual disturbance.  
Moreover, the proposed framework can be easily adopted to other generative policy methods to disambiguate the multi-modal actions.

\begin{table}[t]
\centering
\caption{Performance of different denoising mechanisms under different ET settings.}
\label{tab:exp:q4}
\setlength{\tabcolsep}{6pt}
\renewcommand{\arraystretch}{1.25}
\small
\begin{tabular}{lccc}
    \toprule
    \textbf{Policy} 
    & \textbf{FM} 
    & \textbf{DDIM} \\
    \midrule
    w/o TF & 0.00\%  & 0.00\% \\
    w/ TF & 58.33\% & 66.67\% \\
    \bottomrule
\end{tabular}
\end{table}

\section{Conclusion} 
\label{sec:conclusion}

This paper addresses a fundamental limitation of generative imitation-learning policies in long-horizon robotic manipulation: \emph{multi-modal action ambiguity} (MA$^2$).
This issue is caused by the one-to-many observation–action mapping in long-horizon tasks, where visually indistinguishable observations recur at different execution stages and correspond to distinct valid actions.
Conditioning action generation solely on instantaneous visual inputs often results in ambiguous action selection, leading to frequent task failures.
To resolve this issue, we propose \textbf{Trace-Focused Diffusion Policy (\m{})}, a lightweight and execution-aware framework that explicitly conditions diffusion-based policies on robot history motions. 
By projecting historical execution information into the visual observation space and rendering a trace-focused field, TF-DP provides stage-aware context that disambiguates action selection while emphasizing task-relevant regions to suppress the visual disturbances in the background. 
Extensive real-world experiments on long-horizon manipulation tasks demonstrate that \m{} substantially improves temporal consistency, task success rates, and robustness compared to standard diffusion policies and history-conditioned baselines, while incurring negligible additional computational overhead. 
Moreover, we show that the proposed trace-focused mechanism is compatible with multiple generative policy formulations, highlighting its generality and practicality.

In future work, we will explore extending the trace-focused mechanism to better exploit the rich 3D geometric information inherently encoded in execution traces, which is currently utilized only through a 2D projection in our approach. In addition, while the current trace representation primarily provides coarse task guidance, we plan to investigate its effectiveness in more precise and fine-grained manipulation tasks.

\clearpage
\newpage
\bibliographystyle{assets/plainnat}
\bibliography{paper}

\clearpage
\newpage

\appendix

\section{Appendix}
\section*{Supplementary Materials}
\setcounter{table}{0}
\renewcommand{\thetable}{S\Roman{table}}
\renewcommand*{\thefigure}{S1}
\begin{figure*}
	\centering
	\includegraphics[width=1\linewidth]{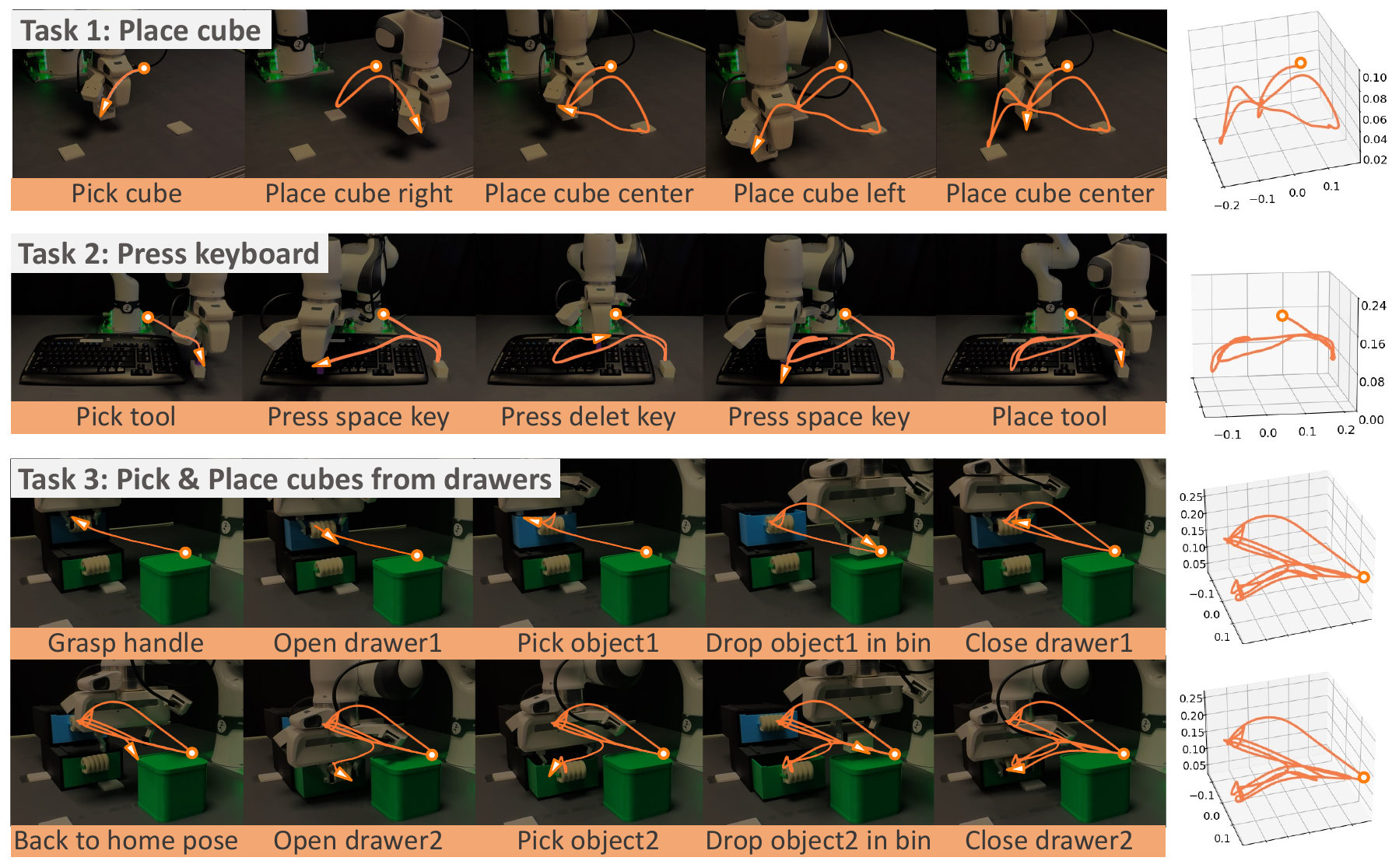}
	\caption{\textbf{Trajectory--action aligned visualization for tasks with MA$^2$.} Representative end-effector trajectories are decomposed into segments corresponding to manually identified manipulation stages for visualization.
Unlike the aggregated trajectories in \fref{fig:exp:traj}, this view highlights how different portions of a long-horizon trajectory correspond to distinct action stages, improving interpretability.}
	\label{fig:traj_action}
\end{figure*}

\subsection{Implementation details}
All experiments are conducted using a unified training and inference configuration across tasks to ensure fair comparison, as summarized in Tab. \ref{tab:training_config}.
At each inference step, the policy conditions on a fixed temporal horizon of eight steps, using the most recent three observations as input and predicting a short sequence of four future actions.
This design balances temporal context and computational efficiency, enabling responsive closed-loop execution while retaining sufficient history for action disambiguation.

For visual observations, all RGB inputs from the wrist, side, and auxiliary cameras are resized to $128\times128$ and jointly used as global conditioning signals.
Low-dimensional proprioceptive inputs, including the end-effector state, are concatenated with visual features to provide complementary geometric information.
During inference, no future action visibility or past action leakage is allowed, ensuring that all decisions rely solely on observable execution history.

All reported results are obtained using a single checkpoint selected at 150 training epochs, which consistently yields the highest success rate across tasks.
Using a fixed checkpoint avoids per-task tuning and reflects the robustness of the learned policy under a common training budget.

\subsection{Trajectory--action alignment analysis}
In the main paper, execution behavior is primarily illustrated using full end-effector trajectories (Fig.~4), which provide a compact summary of long-horizon motion.
However, when visualized as a single continuous curve, it can be difficult to associate specific portions of the trajectory with the underlying manipulation steps being performed. To improve interpretability, we provide an action-aligned visualization that decomposes representative trajectories into segments corresponding to manually identified manipulation stages, such as grasping, intermediate motion, interaction, and returning to a shared workspace region Fig.~\ref{fig:traj_action}.

By aligning trajectory segments with intuitive action stages, the visualization clarifies how similar spatial regions may be visited multiple times during execution while corresponding to different actions at different moments.
Notably, multiple stages revisit visually similar spatial configurations—such as the center placement region in Task~1, the hovering state above the keyboard in Task~2, and the home pose between drawer interactions in Task~3.
This view helps disambiguate the temporal structure of long-horizon trajectories and provides a clearer illustration of the execution process shown in \fref{fig:exp:traj}.

\begin{table}[t]
\centering

\captionsetup{justification=centering}
\caption{Training configuration used for all experiments.}
\begin{tabular}{l l}
\toprule
\textbf{Parameter} & \textbf{Value} \\
\midrule
Optimizer & AdamW \\
Learning rate & $1\times10^{-4}$ \\
AdamW betas & $(0.95,\;0.999)$ \\
AdamW $\epsilon$ & $1\times10^{-8}$ \\
Weight decay & $1\times10^{-6}$ \\
Batch size & 32 \\
Diffusion steps & 100 \\
Noise schedule & Cosine \\
Prediction target & Noise ($\epsilon$) \\
Action steps & 8 \\
Image resolution & $128\times128$ \\
\bottomrule
\end{tabular}
\label{tab:training_config}
\end{table}

\end{document}